# Deep-learning-based upscaling method for geologic models via theory-guided convolutional neural network

Nanzhe Wang[a], Qinzhuo Liao[b], Haibin Chang[a], and Dongxiao Zhang[c,*]

[a] BIC-ESAT, ERE, and SKLTCS, College of Engineering, Peking University, Beijing 100871, P. R. China
[b] CPG, King Fahd University of Petroleum and Minerals, Dhahran, Saudi Arabia
[c] School of Environmental Science and Engineering, Southern University of Science and Technology, Shenzhen 518055, P. R. China

[*]Corresponding author: E-mail address: zhangdx@sustech.edu.cn (Dongxiao Zhang)

**Abstract**

Large-scale or high-resolution geologic models usually comprise a huge number of grid blocks, which can be computationally demanding and time-consuming to solve with numerical simulators. Therefore, it is advantageous to upscale geologic models (e.g., hydraulic conductivity) from fine-scale (high-resolution grids) to coarse-scale systems. Numerical upscaling methods have been proven to be effective and robust for coarsening geologic models, but their efficiency remains to be improved. In this work, a deep-learning-based method is proposed to upscale the fine-scale geologic models, which can assist to improve upscaling efficiency significantly. In the deep learning method, a deep convolutional neural network (CNN) is trained to approximate the relationship between the coarse grid of hydraulic conductivity fields and the hydraulic heads, which can then be utilized to replace the numerical solvers while solving the flow equations for each coarse block. In addition, physical laws (e.g., governing equations and periodic boundary conditions) can also be incorporated into the training process of the deep CNN model, which is termed the theory-guided convolutional neural network (TgCNN). With the physical information considered, dependence on the data volume of training the deep learning models can be reduced greatly. Several subsurface flow cases are introduced to test the performance of the proposed deep-learning-based upscaling method, including 2D and 3D cases, and isotropic and anisotropic cases. The results show that the deep learning method can provide equivalent upscaling accuracy to the numerical method, and efficiency can be improved significantly compared to numerical upscaling.

**Keywords:** upscaling; deep learning; theory-guided convolutional neural network; subsurface flow.

**1 Introduction**

Numerical simulation has become an important tool for investigating subsurface flow problems, such as groundwater flow, carbon capture and storage, oil and gas production, etc. The characterization of large-scale or high-resolution geologic models usually requires a large number of grid blocks in numerical simulation, and it would be computationally expensive and time-consuming to run the fine-scale models directly with numerical simulators. For tasks that necessitate simulating multiple realizations of geologic models or running the simulation iteratively, e.g., uncertainty quantification and data assimilation, the computational burden would be further increased. Upscaling the geologic models from fine-scale to coarse-scale can assist to reduce the requisite computational effort for running the models, which has constituted an essential and necessary task for subsurface flow simulation (Chen et al., 2003; Dagan et al., 2013; Durlofsky, 1991; Li & Durlofsky, 2016; Liao et al., 2019; Wen & Gómez-Hernández, 1996).

Upscaling of fine-scale hydraulic conductivity requires calculation of equivalent conductivity over the coarse block consisting of high-resolution grid blocks, and the simulated hydraulic heads and velocity with the upscaled conductivity should match the reference fine-scale solutions as closely as possible. There are many upscaling techniques in the existing literature, and systematic reviews of upscaling methods can be found in previous studies (Dagan et al., 2013; Farmer, 2002; Renard & de Marsily, 1997; Wen & Gómez-Hernández, 1996). Two categories of upscaling methods are mainly introduced here, including analytical methods and numerical methods (Liao et al., 2019; Renard & de Marsily, 1997). Analytical methods are both straightforward and efficient, such as arithmetic average, geometric average, harmonic average, arithmetic-harmonic average, and harmonic-arithmetic average (Durlofsky, 1992; Wen & Gómez-Hernández, 1996). Cardwell and Parsons (1945) proved that the block equivalent conductivity is bounded by the harmonic-arithmetic average and the arithmetic-harmonic average of fine-scale grids in the block, which is termed Cardwell and Parsons bounds, and Guerillot et al. (1990) proposed to calculate the geometric average between the two bounds. Journel et al. (1986) reported a power average to calculate equivalent conductivity for coarse blocks. With the exponent value varying from -1 to 1, the calculated equivalent conductivity can vary between the harmonic and the arithmetic averages. Moreover, the determination of the exponent value is case-specific, and depends on the type of heterogeneity, the block shape and size, and flow conditions (Wen & Gómez-Hernández, 1996). Although the analytical methods are simple and fast, accuracy and robustness are limited by assumptions associated with the analytical methods. For instance, the above-mentioned methods assume that the conductivity of the grids in the coarse block should be scalars, and the calculated equivalent conductivity for the block is also assumed to be a scalar. Liao et al. (2019) developed an analytical upscaling method based on perturbation expansion techniques and Fourier analysis, which can provide accurate estimations for equivalent conductivity in heterogeneous and anisotropic 2D cases. The analytical method was also further extended to 3D scenarios (Liao et al., 2020).

Numerical upscaling methods are relatively more accurate and robust for heterogeneous and anisotropic subsurface flow problems. In numerical methods, the flow equations need to be solved for each coarse block, and the finite difference (FD) method is usually utilized. While solving the flow equations for the coarse blocks, the boundary conditions imposed on the blocks would also affect the upscaled equivalent conductivity. The commonly used boundary conditions include fixed head-no flow boundary conditions, fixed head boundary conditions on all sides, and periodic boundary conditions (Durlofsky, 1991; Liao et al., 2019; Wen et al., 2003; Zhou et al., 2010). White and Horne (1987) reported a numerical upscaling algorithm, in which a set of different boundary conditions are imposed, and the simulation results are averaged to calculate the coarse equivalent conductivity. Durlofsky (1991) utilized periodic boundary conditions for numerical calculation of equivalent conductivity tensors, which can provide symmetric equivalent conductivity tensors. Therefore, upscaled equivalent conductivity with periodic boundary conditions seems to possess more desirable features. Even though numerical methods can provide more robust and accurate estimation of equivalent coarse-scale conductivity, computational burden remains a major challenge, especially for large-scale cases with millions of grid blocks, because the flow equations need to be solved numerically and repeatedly for each coarse block.

Analogous tasks to geologic model upscaling also exist in the deep learning field. For example, image super-resolution tasks in computer vision require recovering a high-resolution image from the low-resolution original image, which can be regarded as an inverse process of upscaling (Dong et al., 2014, 2016; Ledig et al., 2017). Inspired by super-resolution tasks in computer vision, we attempt to utilize deep learning algorithms to achieve upscaling of geologic models. The most straightforward idea is to learn an end-to-end mapping from the fine-scale conductivity fields to the upscaled coarse-scale fields with convolutional neural networks (CNNs). However, this idea derives from a computer vision perspective, rather than a subsurface flow perspective. Specifically, physical principles are not involved in the whole process, and the upscaled coarse equivalent conductivity fields may not produce hydraulic head predictions that match the reference fine-scale solutions. Furthermore, a large number of training data are required to train such a deep learning model, which is computationally prohibitive.

In this work, instead of training a global mapping directly, a more efficient and robust deep learning upscaling method is proposed, in which the lightweight CNNs can be trained to predict the solutions of flow equations for local coarse-scale blocks, and the numerical solving process can be replaced with the trained models. The equivalent block conductivity can then be calculated following the numerical procedure. Rather than solving the flow equations for each coarse block numerically, the trained CNN models can predict the solutions instantly and in parallel, and thus the efficiency of upscaling can be improved significantly. Because the constructed CNN models are lightweight and only applied to the local blocks, the volume of requisite training data is relatively small. In addition, the flow equations and boundary conditions can also be incorporated into the training process as prior physical knowledge to regulate the CNN models, which is termed the theory-guided convolutional neural network (TgCNN) (Wang et al., 2021b). With the theory-guidance, the dependence of the data volume can be further reduced, and the predictions from the TgCNN can be forced to follow physical laws. Several subsurface flow cases are introduced to test the performance of the proposed deep-learning-based upscaling method, including 2D and 3D cases, and isotropic and anisotropic cases. The results demonstrate that the proposed method can provide equivalent upscaling accuracy to the numerical method, and efficiency can be improved markedly compared to numerical upscaling.

The remainder of this paper is organized as follows. In section 2, the governing equations of single-phase subsurface flow and numerical upscaling method are introduced, and the proposed deep-learning-based method is also illustrated. In section 3, several cases are designed to test the performance of the proposed method, including 2D and 3D cases, and isotropic and anisotropic cases. In section 4, discussions and conclusions are provided.

## 2 Methodology

### 2.1 Governing equations

In this work, incompressible single-phase steady-state subsurface flow problems are studied, which are governed by the following equation:

$$-\nabla \cdot (\mathbf{K}(\mathbf{x})\nabla H(\mathbf{x})) = 0, \quad \mathbf{x} \in \Omega \tag{1}$$

and subjected to the following boundary conditions:

$$\begin{cases} H(\mathbf{x}) = f(\mathbf{x}), & \mathbf{x} \in \Omega_D \\ \mathbf{K}(\mathbf{x})\nabla H(\mathbf{x}) \cdot \mathbf{n}(\mathbf{x}) = g(\mathbf{x}), & \mathbf{x} \in \Omega_N \end{cases} \tag{2}$$

where $\mathbf{K}$ denotes the hydraulic conductivity tensor $[LT^{-1}]$; $H$ denotes the hydraulic head $[L]$; $\Omega \subset R^n$ denotes the studied domain with Dirichlet boundary $\Omega_D$ and Neumann boundary $\Omega_N$; $f(\mathbf{x})$ denotes the prescribed head on Dirichlet boundary segments $[L]$; and

$g(\mathbf{x})$ denotes the prescribed flux across Neumann boundary segments $[LT^{-1}]$.

By taking the average flow rate over the domain, the equivalent hydraulic conductivity for the domain can be defined with the following equation (Renard & de Marsily, 1997; Rubin & Gómez-Hernández, 1990):

$$\frac{1}{V}\int_\Omega \mathbf{v}(\mathbf{x})d\mathbf{x} = \mathbf{K}_{eq}\left(\frac{1}{V}\int_\Omega \nabla H(\mathbf{x})d\mathbf{x}\right) \quad (3)$$

where $V$ denotes the volume of the domain; $\mathbf{v}$ denotes the Darcy filtration velocity $[LT^{-1}]$, which can be calculated with $\mathbf{v} = -\mathbf{K}\nabla H$ (Darcy's law); and $\mathbf{K}_{eq}$ denotes the equivalent hydraulic conductivity of the domain $\Omega$ $[LT^{-1}]$. The hydraulic conductivity of the domain to be upscaled is assumed to be diagonal, as represented below:

$$\mathbf{K} = \begin{pmatrix} Kx & \\ & Ky \end{pmatrix} \text{ (2D) and } \mathbf{K} = \begin{pmatrix} Kx & & \\ & Ky & \\ & & Kz \end{pmatrix} \text{ (3D),} \quad (4)$$

and the upscaled equivalent conductivity can be represented with a tensor:

$$\mathbf{K}_{eq} = \begin{pmatrix} \bar{K}xx & \bar{K}xy \\ \bar{K}yx & \bar{K}yy \end{pmatrix} \text{ (2D) and } \mathbf{K}_{eq} = \begin{pmatrix} \bar{K}xx & \bar{K}xy & \bar{K}xz \\ \bar{K}yx & \bar{K}yy & \bar{K}yz \\ \bar{K}zx & \bar{K}zy & \bar{K}zz \end{pmatrix} \text{ (3D).} \quad (5)$$

## 2.2 Heterogeneity

In subsurface flow problems, geologic models are usually heterogeneous. In this work, the Karhunen–Loeve expansion (KLE) is utilized to characterize and generate the heterogeneous and spatially correlated hydraulic conductivity fields (Ghanem & Spanos, 2003; Zhang & Lu, 2004). The log-transformed hydraulic conductivity $\ln K(\mathbf{x}, \omega)$ can be seen as a stochastic field (represented as $Y(\mathbf{x}, \omega) = \ln K(\mathbf{x}, \omega)$ for simplicity), which can be expressed in the following form using the KLE (Ghanem & Spanos, 2003):

$$Y(\mathbf{x}, \omega) = \langle Y(\mathbf{x}, \omega)\rangle + \sum_{i=1}^{\infty} \sqrt{\lambda_i} f_i(\mathbf{x}) \xi_i(\omega), \quad (6)$$

where $\langle Y(\mathbf{x}, \omega)\rangle$ denotes the mean of the stochastic field; $\lambda_i$ and $f_i(\mathbf{x})$ are the eigenvalue and eigenfunction of the covariance, respectively; and $\xi_i(\omega)$ denotes the independent standard Gaussian random variables when $Y(\mathbf{x},\omega)$ is a Gaussian random field. Considering that there are infinite terms in Eq. (6), the expansion can be truncated into finite terms ($n$) according to the decay rate of $\xi_i(\omega)$. The truncated number $n$ can be determined by calculating the preserved percentage of energy (information of the stochastic field) $\sum_{i=1}^{n}\lambda_i \Big/ \sum_{i=1}^{\infty}\lambda_i$. If the retained energy reaches the predefined weight, the expansion can then be truncated, and the stochastic field $Y(\mathbf{x},\omega) = \ln K(\mathbf{x},\omega)$ can be parameterized with the $n$ dimensional random vector:

$$\boldsymbol{\xi} = \{\xi_1(\omega), \xi_2(\omega), \cdots, \xi_n(\omega)\}. \tag{7}$$

Therefore, the stochastic field can be represented as:

$$Y(\mathbf{x},\omega) = \langle Y(\mathbf{x},\omega) \rangle + \sum_{i=1}^{n}\sqrt{\lambda_i} f_i(\mathbf{x})\xi_i(\omega), \tag{8}$$

and the realizations of hydraulic conductivity can be generated by sampling the independent random variables $\xi_i(\omega)$. Each group of random variables corresponds to a realization of conductivity field.

## 2.3 Numerical methods

Numerical methods have been widely used for upscaling of fine-scale geologic models (Durlofsky, 1991). In numerical methods, the fluid flow equation Eq. (1) can be solved for each coarse grid block. Moreover, periodic boundary conditions can be exerted on the coarse grid blocks while solving the flow equations (Durlofsky, 1991), as presented below (2-dimensional scenarios):

$$\begin{aligned} H(x,y)\big|_{x=0} &= H(x,y)\big|_{x=L_x} - \Delta H_1, \quad v_x(x,y)\big|_{x=0} = v_x(x,y)\big|_{x=L_x} \\ H(x,y)\big|_{y=0} &= H(x,y)\big|_{y=L_y} - \Delta H_2, \quad v_y(x,y)\big|_{y=0} = v_y(x,y)\big|_{y=L_y} \end{aligned} \tag{9}$$

where $L_x$ and $L_y$ denote the length of the coarse-grids in the x- and y-direction, respectively; $\Delta H_1$ and $\Delta H_2$ denote the constant hydraulic head differences between the boundaries; and

$v_x$ and $v_y$ denote the Darcy velocity in the x- and y-direction, respectively $[LT^{-1}]$.

To solve flow equations for different coarse grid blocks, the finite difference (FD) method can be adopted, in which the governing equation can be discretized as follows (2D scenarios):

$$\frac{Kx_{i+1/2,j}(H_{i+1,j}-H_{i,j})-Kx_{i-1/2,j}(H_{i,j}-H_{i-1,j})}{(\Delta x)^2}+\frac{Ky_{i,j+1/2}(H_{i,j+1}-H_{i,j})-Ky_{i,j-1/2}(H_{i,j}-H_{i,j-1})}{(\Delta y)^2}=0$$

(10)

where $i$ and $j$ denote the index of grid blocks in the x- and y-direction, respectively; $Kx$ and $Ky$ denote the hydraulic conductivity component in the x- and y-direction, respectively; and $Kx_{i+1/2,j}$ denotes the conductivity at the interface of grid block $(i, j)$ and $(i+1, j)$, which can be calculated with the harmonic average of two conductivities at two adjacent grid blocks as follows:

$$Kx_{i+1/2,j}=\frac{2}{1/Kx_{i,j}+1/Kx_{i+1,j}} \quad (11)$$

By introducing transmissibility, Eq. (10) can be further rewritten as:

$$\frac{Tx_{i+1/2,j}(H_{i+1,j}-H_{i,j})}{\Delta x}-\frac{Tx_{i-1/2,j}(H_{i,j}-H_{i-1,j})}{\Delta x}+\frac{Ty_{i,j+1/2}(H_{i,j+1}-H_{i,j})}{\Delta y}-\frac{Ty_{i,j-1/2}(H_{i,j}-H_{i,j-1})}{\Delta y}=0$$

(12)

where $Tx$ and $Ty$ denote the transmissibility between the two adjacent grid blocks, which can be defined as:

$$Tx_{i+1/2,j}=\frac{2\Delta y}{1/Kx_{i,j}+1/Kx_{i+1,j}}, \quad (13)$$

$$Ty_{i,j+1/2}=\frac{2\Delta x}{1/Ky_{i,j}+1/Ky_{i,j+1}}. \quad (14)$$

The linear system of equations can then be constructed based on Eq. (12) for the coarse grid blocks, and the sparse coefficient matrix can also be assembled for solving the linear equations. With the solutions of hydraulic heads, the equivalent hydraulic conductivity of the coarse grid blocks can be calculated by Eq. (3). The workflow of the numerical upscaling method is presented in **Figure 1**.

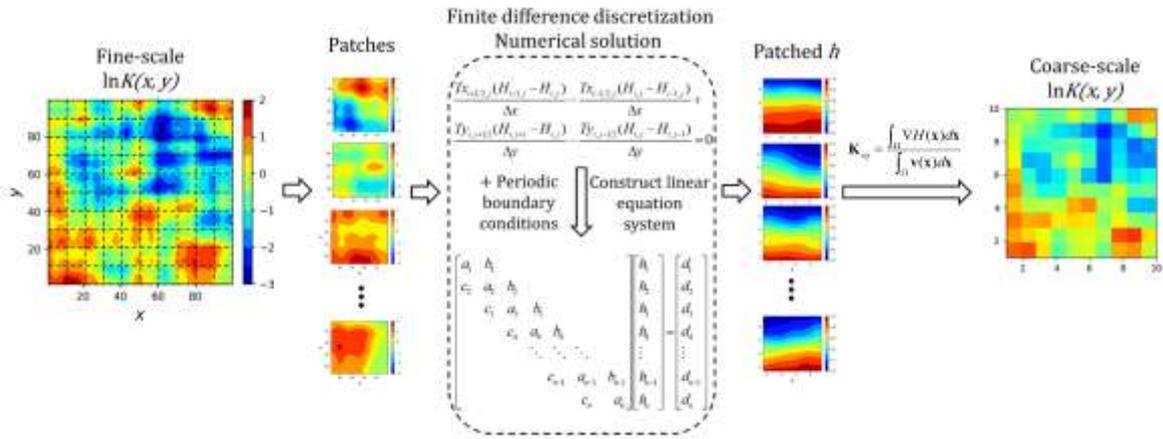

**Figure 1.** Workflow of the numerical upscaling method.

It is worth noting that for each coarse grid block, the periodic boundary conditions should be exerted twice (or three times for 3D scenarios) with different directions, and the flow equations should also be solved twice accordingly to obtain the equivalent conductivity with different principal directions, as presented in **Figure 2**.

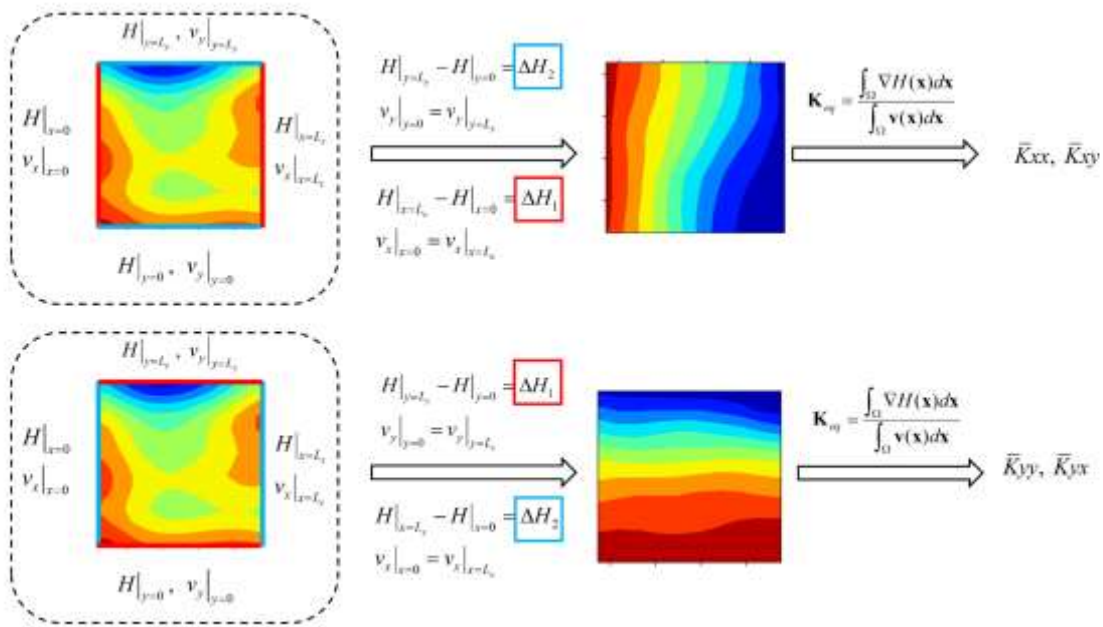

**Figure 2.** Illustration of periodic boundary conditions in different directions.

## 2.4 Deep-learning-based methods

The most time-consuming part of numerical upscaling methods is solving the flow equations for each coarse block. Therefore, in order to improve the efficiency of the upscaling process, numerical solving for each coarse grid can be replaced by deep learning models, as shown in **Figure 3**. The deep learning model can approximate the mapping from the coarse patch of the fine-scale hydraulic conductivity field to the local hydraulic head distribution, which can assist to avoid cumbersome numerical solving and provide a more efficient forward calculation. The deep learning model can be represented as follows:

$$\hat{H} = g(K;\ \theta), \tag{15}$$

where $\hat{H}$ denotes the predicted hydraulic head; and $g(\theta)$ denotes the deep learning model with model parameters $\theta$ (weights and bias).

Considering that the patch of the fine-scale hydraulic conductivity field and the distribution of hydraulic heads can both be seen as images, the convolutional neural network (CNN) structure can be adopted to approximate the relationship, which has constituted a powerful tool for image processing (Goodfellow et al., 2016; He et al., 2016; Ledig et al., 2017). A convolutional encoder-decoder architecture can be constructed easily with several convolution layers and deconvolution layers. The most straightforward approach to train the CNN model is using the data pairs obtained from numerical solutions. The loss function can be written as follows:

$$L_{Data}(\theta) = \frac{1}{N_{grid}} \frac{1}{N} \sum_{i=1}^{N} \left\| \hat{H}_i - H_i \right\|_2^2 = \frac{1}{N_{grid}} \frac{1}{N} \sum_{i=1}^{N} \left\| g(K_i;\theta) - H_i \right\|_2^2, \tag{16}$$

where $N$ denotes the total number of data pairs (image-to-image pairs between the hydraulic conductivity and heads); $N_{grid}$ denotes the number of grids in each realization; $\hat{H}_i$ denotes the images of predicted hydraulic heads for hydraulic conductivity $K_i$; and $H_i$ denotes the ground truth of hydraulic heads for the $i$th data pair.

When the available data pairs are abundant, minimizing the loss function Eq. (16) would be the most direct and convenient way to train the deep CNNs. It would be challenging, however, to train an accurate CNN model in scenarios with limited labeled data pairs. Under this circumstance, already known physical laws and governing equations could provide some additional and valuable information for the training stage to construct the theory-guided convolutional neural network (TgCNN). By incorporating physical laws or equations as prior knowledge to guide the training process, the problem of training data shortage can be effectively alleviated. In order to achieve theory-guided training, consider the discretized governing equation Eq. (12), which can be utilized to regulate CNNs in the training process. The predictions of hydraulic heads shall honor Eq. (12), from which the ground truth comes, and thus when the predicted values of hydraulic heads are substituted into Eq. (12), the residuals of the equation should approach zero, as presented in Eq. (17):

$$R(K;\theta) = \frac{Tx_{i+1/2,j}(\hat{H}_{i+1,j} - \hat{H}_{i,j})}{\Delta x} - \frac{Tx_{i-1/2,j}(\hat{H}_{i,j} - \hat{H}_{i-1,j})}{\Delta x} \\ + \frac{Ty_{i,j+1/2}(\hat{H}_{i,j+1} - \hat{H}_{i,j})}{\Delta y} - \frac{Ty_{i,j-1/2}(\hat{H}_{i,j} - \hat{H}_{i,j-1})}{\Delta y} \to 0, \qquad (17)$$

Therefore, the residuals can be minimized in the training process to force the CNN model to produce predictions that adhere to governing equation Eq. (12). The residual term in the loss function can be expressed as follows:

$$L_{GE}(\theta) = \frac{1}{N_{grid}} \frac{1}{N_r} \sum_{i=1}^{N_r} \|R(K_i;\theta)\|_2^2, \qquad (18)$$

where $N_r$ denotes the total number of hydraulic conductivity patches used to calculate the equation residuals and boundary condition residuals, which is different from the number of paired training data $N$. It is worth mentioning that the labeled data are not required while calculating the residuals, and only the conductivity patches are utilized.

Since the periodic boundary conditions need to be exerted while solving the flow equations for each coarse grid block, the periodic boundary conditions can also be incorporated as prior knowledge into the training stage. The loss terms for the boundary conditions can be written as:

$$L_{BC-H}(\theta) = \frac{1}{N_{grid-b}} \frac{1}{N_r} \sum_{i=1}^{N_r} \left\| \left(\hat{H}\big|_{x=0} - \hat{H}\big|_{x=L_x}\right) - \Delta H_1 \right\|_2^2 + \frac{1}{N_{grid-b}} \frac{1}{N_r} \sum_{i=1}^{N_r} \left\| \left(\hat{H}\big|_{y=0} - \hat{H}\big|_{y=L_y}\right) - \Delta H_2 \right\|_2^2,$$

$$L_{BC-v}(\theta) = \frac{1}{N_{grid-b}} \frac{1}{N_r} \sum_{i=1}^{N_r} \left\| \hat{v}_x \big|_{x=0} - \hat{v}_x \big|_{x=L_x} \right\|_2^2 + \frac{1}{N_{grid-b}} \frac{1}{N_r} \sum_{i=1}^{N_r} \left\| \hat{v}_y \big|_{y=0} - \hat{v}_y \big|_{y=L_y} \right\|_2^2, \quad (20)$$

where $N_{grid-b}$ denotes the number of grids at boundaries; and $\hat{v}_x$ and $\hat{v}_y$ can be calculated, respectively, using Darcy's law as follows:

$$\hat{v}_x = -K_x \cdot \nabla \hat{H}, \quad (21)$$

$$\hat{v}_y = -K_y \cdot \nabla \hat{H}, \quad (22)$$

Consequently, the loss function of TgCNN after incorporating the theory-guidance can be given as:

$$L(\theta) = \lambda_{Data} L_{Data}(\theta) + \lambda_{GE} L_{GE}(\theta) + \lambda_{BC-H} L_{BC-H}(\theta) + \lambda_{BC-v} L_{BC-v}(\theta). \quad (23)$$

where $\lambda_{Data}$, $\lambda_{GE}$, $\lambda_{BC-H}$, and $\lambda_{BC-v}$ denote the weights of different loss terms in the total loss function. These hyperparameters can be set as 1 by default, and could also be further adjusted by balancing the magnitudes of each loss term.

The TgCNN model can be trained by minimizing the loss function Eq. (23), which can then be used to predict the hydraulic head solutions for different conductivity patches. The equivalent hydraulic conductivity at the coarse scale can be further calculated with the predicted hydraulic head, as shown in **Figure 3**. With the numerical solution process being replaced by the deep learning models, the efficiency of upscaling geologic models can be improved significantly.

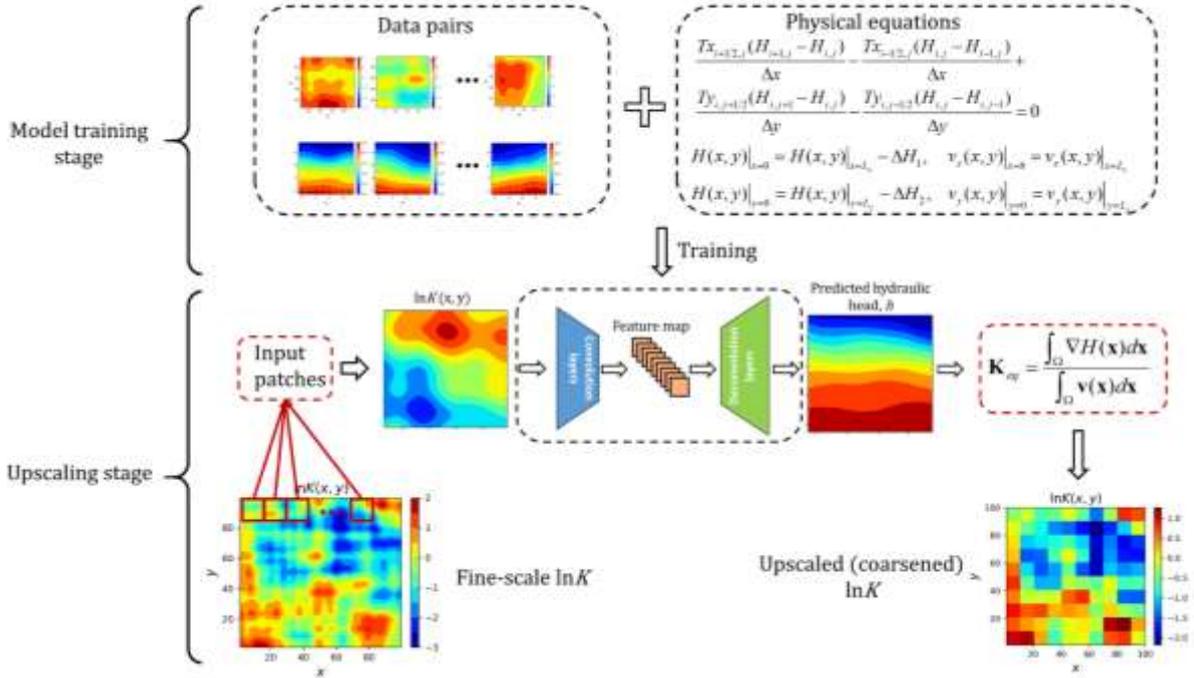

**Figure 3.** Structure and workflow of the deep-learning-based upscaling method.

**3 Case Studies**

In this subsection, several synthetic cases are introduced to test the performance of the proposed deep-learning-based upscaling method, including 2D cases and 3D cases.

**3.1 2D case**

*3.1.1 2D base case*

A 2D case is firstly studied in this subsection. The physical domain is a square area, with each side of length 100 [L] ($L_x = L_y = 100$ [L]). The correlation length of the x- and y-direction are both set to be 20 [L] ($\eta_x = \eta_y = 20$ [L]). The mean of the stochastic log-transformed hydraulic conductivity field is 0, and the variance is 1 ($E(\ln K) = 0$ and $\sigma^2_{\ln K} = 1.0$). In addition, the covariance of the stochastic field is an exponential function:

$$C_{\ln K}(\mathbf{x}, \mathbf{x}') = \sigma^2_{\ln K} \exp(-\frac{|x_1 - x_2|}{\eta_x} - \frac{|y_1 - y_2|}{\eta_y}). \tag{24}$$

The realizations of hydraulic conductivity fields can then be generated with KLE. In order to characterize the fields more elaborately, 90% of the information of the stochastic fields is preserved while generating the hydraulic conductivity for the fine-scale grid blocks. In this case, the hydraulic conductivity is assumed to be isotropic, i.e., $Kx = Ky$.

The fine-scale geologic models are $100 \times 100$ grid blocks, which need to be upscaled into $10 \times 10$ coarse grid blocks, with each coarse block consisting of $10 \times 10$ fine-scale blocks. In the numerical upscaling method, the flow equations need to be solved for each coarse grid block with the periodic boundary conditions. In this case, the constant hydraulic head difference of the two directions are set to be $\Delta H_1 = 1$ and $\Delta H_2 = 0$, respectively. In the deep-learning-based upscaling method, the TgCNN model is constructed to predict the solutions of flow equations and improve the efficiency of the upscaling process. Considering that each coarse block consists of $10 \times 10$ high-resolution grids, the image size of inputs to the TgCNN model is also $10 \times 10$. Moreover, the size of outputted hydraulic head images is $12 \times 12$ because the heads at the surface of domain boundaries are also incorporated. The details of the TgCNN structure are presented in **Table 1**. To train the TgCNN model, five fine-scale hydraulic conductivity fields are generated with KLE and divided into 500 patches (coarse grid blocks). The 500 patches of hydraulic conductivity are utilized to calculate the equation residuals and impose the physical constraints. It is worth noting that no labeled data are used to train the TgCNN model, and only the physical laws are exploited to provide the valuable information for training. Therefore, the loss function of the TgCNN model in this case can be rewritten as:

$$L(\theta) = \lambda_{GE} L_{GE}(\theta) + \lambda_{BC-H} L_{BC-H}(\theta) + \lambda_{BC-v} L_{BC-v}(\theta). \tag{25}$$

As previously mentioned, the theory-guidance can assist to alleviate the shortage of labeled datasets for the training of deep learning models, and the label-free TgCNN model in this case can prove this again. The weights of different terms in the loss function are set to be 1 in this case. The learning rate is set to be 0.001 initially, and it decays 10% after each 100 epochs. It takes approximately 158.657 s to finish the training of the model (1,000 epochs) on an NVIDIA TITAN RTX GPU. It can be seen that it is both convenient and fast to train such a lightweight deep learning model.

Table 1. Architecture of the TgCNN model for the 2D case.

| | Layers | Output size | Number of channels |
|---|---|---|---|
| | Input | 10 × 10 | 1 |
| Encoder | Convolution (k3s1p1) | 10 × 10 | 16 |
| | Activation (Swish) | 10 × 10 | 16 |
| | Convolution (k3s1p1) | 10 × 10 | 32 |
| | Activation (Swish) | 10 × 10 | 32 |
| | Convolution (k3s1p0) | 8 × 8 | 64 |
| | Activation (Swish) | 8 × 8 | 64 |
| | Convolution (k3s1p0) | 6 × 6 | 128 |
| | Activation (Swish) | 6 × 6 | 128 |
| | Convolution (k3s1p0) | 4×4 | 256 |
| | Activation (Swish) | 4×4 | 256 |
| Decoder | Deconvolution (k3s1p0) | 6 × 6 | 128 |
| | Activation (Swish) | 6 × 6 | 128 |
| | Deconvolution (k3s1p0) | 8 × 8 | 64 |
| | Activation (Swish) | 8 × 8 | 64 |
| | Deconvolution (k3s1p0) | 10 × 10 | 32 |
| | Activation (Swish) | 10 × 10 | 32 |
| | Deconvolution (k3s1p0) | 12 × 12 | 16 |
| | Activation (Swish) | 12 × 12 | 16 |
| | Deconvolution (k3s1p1) | 12 × 12 | 1 |
| | Activation (Swish) | 12 × 12 | 1 |

The accuracy of the predicted hydraulic heads from the trained TgCNN model can be examined with numerical solutions. For 100 newly generated hydraulic conductivity patches, the $R^2$ scores (also known as coefficients of determination) between the predictions and the numerical solutions of hydraulic heads are presented in the histogram in **Figure 4**. The scatterplot of hydraulic heads at three different points are shown in **Figure 5**. It can be seen that the constructed TgCNN model can predict the hydraulic head solutions accurately for different hydraulic head patches, even when trained without any labeled data.

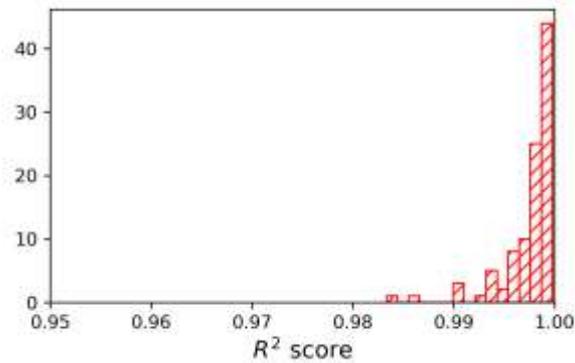

**Figure 4.** Histogram of $R^2$ score between the predictions from the TgCNN model and numerical solutions of hydraulic heads.

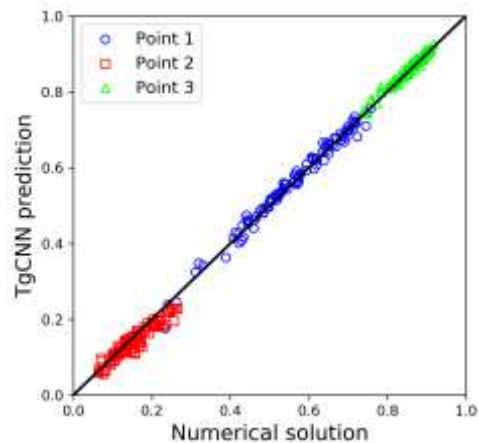

**Figure 5.** Scatterplot of hydraulic heads at three sampled points (Point 1: x=6, y=6; Point 2: x=10, y=10; Point 3: x=3, y=3).

The TgCNN model can then be used for efficient upscaling of fine-scale geologic models following the workflow presented in **Figure 3**. Consider the fine-scale hydraulic conductivity field in **Figure 6** (a) (100×100), which is generated with KLE, and the numerical solution of the flow equations for this realization can be implemented with the MATLAB Reservoir Simulation Toolbox (MRST) (Lie, 2019). **Figure 6** (b), (c), and (d) present the fine-scale hydraulic head, velocity in the x-direction, and streamline, respectively. The upscaled hydraulic head and velocity can then be calculated by averaging the fine-scale solutions directly, which can be regarded as the true coarse-scale solutions or benchmarks, as shown in **Figure 6** (e), (f), and (g). **Figure 6** (h) illustrates the upscaled hydraulic conductivity field with the numerical method following the workflow in **Figure 1**. The upscaled hydraulic head, velocity, and streamline of the numerical method shown in **Figure 6** (i), (j), and (k) can be obtained by inputting **Figure 6** (h) into the MRST solver. **Figure 6** (l), (m), (n), and (o) show the upscaled hydraulic conductivity field, hydraulic head, velocity, and streamline with the proposed deep-learning-based method, respectively. It can be seen that the upscaled conductivity, hydraulic head, velocity and streamline with the deep-learning-based method are almost the same as the numerical results, which demonstrates that the deep learning method can provide an accurate alternative approach for upscaling. Furthermore, both the results from the deep learning method and the numerical method are similar to the benchmarks (true coarse-scale solutions, **Figure 6** (e), (f), and (g)).

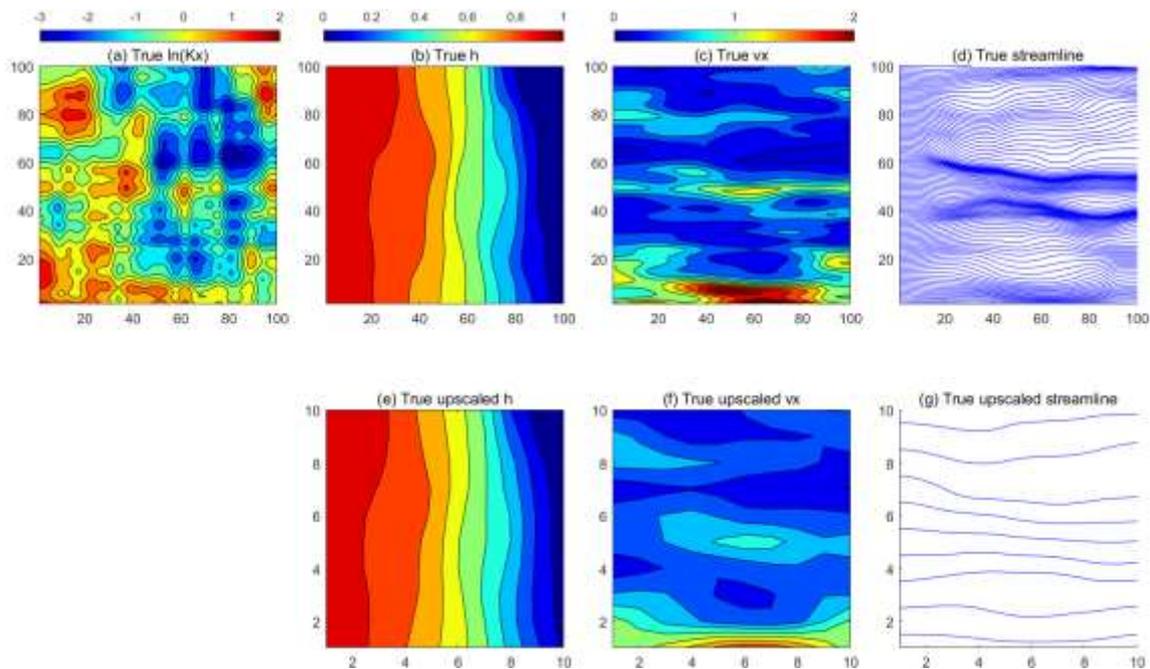

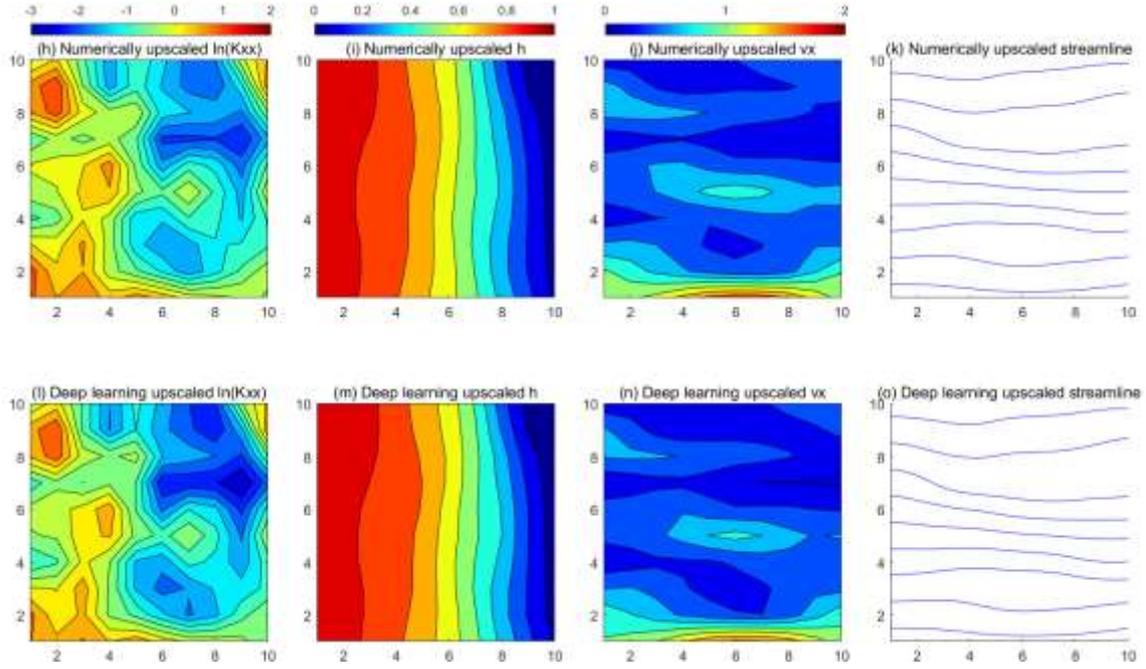

**Figure 6.** Comparison of fine-scale and upscaled log-transformed hydraulic conductivity field, hydraulic head, velocity, and streamline for the 2D case.

The scattered correlation plots of the upscaled results are presented in **Figure 7**. It can be seen that the upscaled hydraulic head and velocity in the x-direction with both the numerical method and the deep learning method can match the true coarse-scale results well. The velocity in the y-direction matches the benchmarks slightly worse, which may be ascribed to the fact that the y-direction is not the principal flow direction ($\Delta H_2 = 0$). The upscaled results with the deep learning method are almost identical to those obtained from the numerical method, as shown in **Figure 7**, which again demonstrates the effectiveness of the deep learning upscaling method. In order to evaluate the performance of the deep learning method statistically, 1,000 fine-scale hydraulic conductivity realizations are generated with KLE, and upscaled with the numerical method and the deep learning method, respectively. The upscaled results of the 1,000 realizations at a sampled point in the domain are scattered in the correlation plots shown in **Figure 8**. It can be seen that the deep learning method can provide satisfactory upscaling performance statistically. Moreover, the efficiency of the proposed deep learning method can also be assessed. **Figure 9** compares the consumed time for upscaling different numbers of fine-scale hydraulic conductivity realizations with different methods. One can see that even though additional time is required to train the TgCNN model, the trained TgCNN model can improve the efficiency of the upscaling process significantly compared to the numerical method, and it takes just several seconds to finish the upscaling of thousands of fine-scale realizations. Therefore, the deep learning method possesses a major advantage in terms of efficiency without the loss of accuracy, and this advantage becomes increasingly obvious as the number of realizations increases.

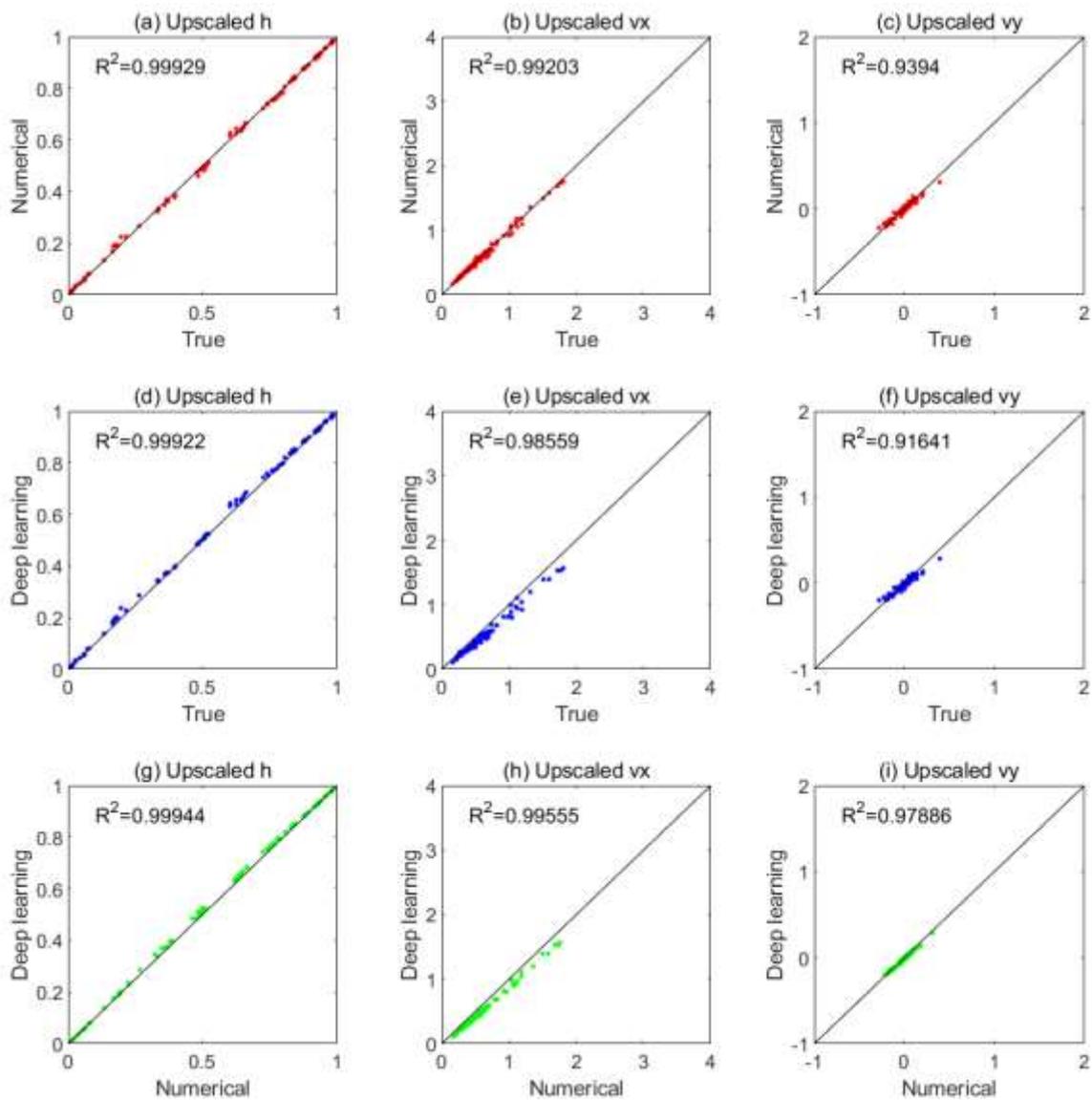

**Figure 7.** Scatterplots of the upscaled results with different methods for the 2D case.

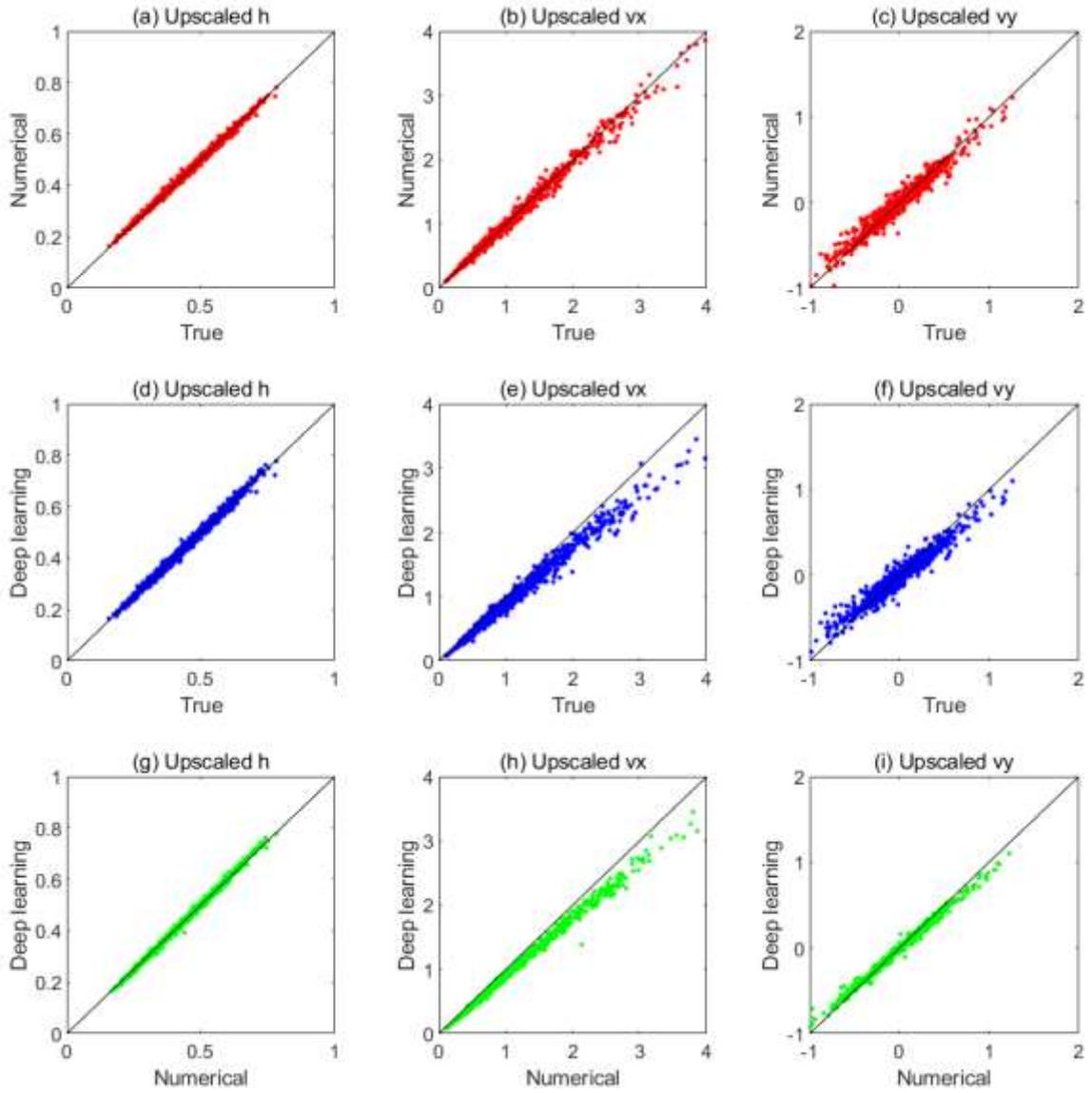

**Figure 8.** Scatterplots of the upscaled results at sampled points (x=6, y=6) for the 1,000 fine-scale realizations.

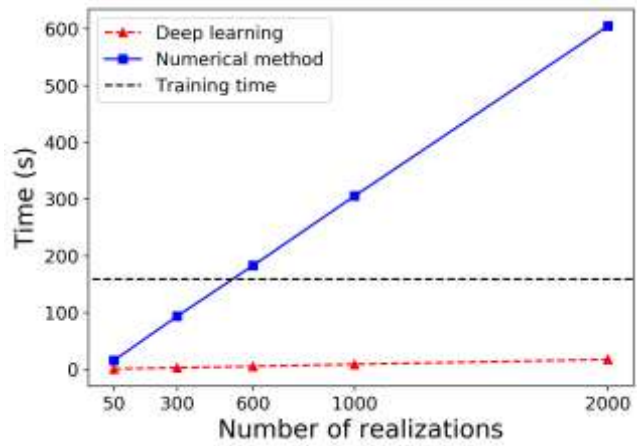

**Figure 9.** Efficiency comparison between the numerical and deep learning upscaling methods.

*3.1.2 The effect of training data*

The TgCNN model in the former base case is trained in a label-free manner, i.e., trained without labeled training data and only with physical constraints. However, the effect of training data on the upscaling accuracy still needs to be investigated. Several CNN models trained with different amounts of training data and without the physical equation terms are utilized for upscaling of 300 fine-scale hydraulic conductivity fields, the accuracy comparison of which is presented in **Figure 10**. The dashed lines in **Figure 10** show the mean of the 300 $R^2$ scores for the upscaled results with CNN models, and it is obvious that the upscaled results become more accurate as the number of training data increases. It can also be seen that when the labeled training data are adequate, the trained CNN model can also provide satisfactory upscaling performance, even when trained without physical constraints. If the labeled training data are limited or computational prohibitive to obtain, however, the guidance of physical laws in the training process could still be beneficial.

In order to elucidate the effect of physical constraints in the training process, several TgCNN models are also trained with different numbers of training data. Similar to the base case in section 3.1.1, 500 coarse block patches are utilized to calculate the equation residuals, and to impose the physical constraints. The comparison of the $R^2$ score average for 300 fine-scale realizations are presented in **Figure 10**. One can see that with the assistance of theory-guidance, the upscaling performance of TgCNN models is markedly more robust, and the accuracy is higher than that using the purely data-driven CNN models. Moreover, the impact of training data is no longer obvious when the physical equations are utilized to constrain the models, which demonstrates the effectiveness of the theory-guided training. In addition, the dependence on the training data volume of the deep learning upscaling framework can be reduced.

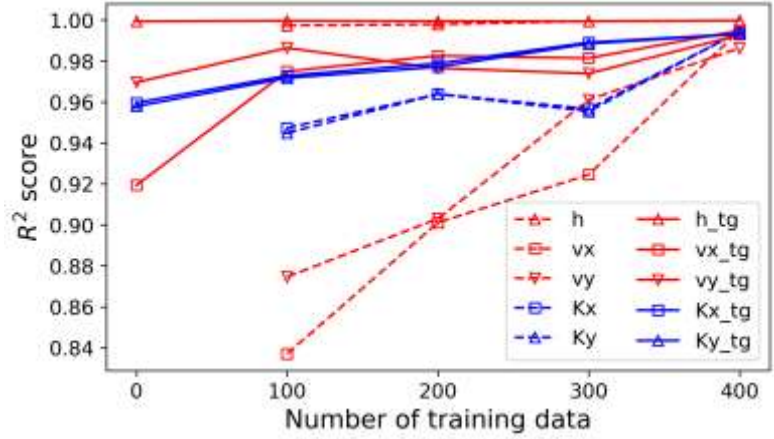

**Figure 10.** $R^2$ score of the upscaled results from the deep learning method with different numbers of training data.

### *3.1.3 The effect of upscaling ratio*

In this subsection, the performance of the deep learning method with different upscaling ratios is investigated. Still consider the base case in section 3.1.1. To upscale the fine-scale conductivity field with upscaling ratios 5 and 20, two TgCNN models are trained with inputting image size $5\times5$ and $20\times20$, respectively. The upscaled results for the fine-scale realizations in section 3.1.1 with upscaling ratios 5 and 20 are presented in **Figure 11** and **Figure 12**, respectively. It is obvious that the upscaled results become more similar to the fine-scale results when the upscaling ratio decreased to 5, because more details of the high-resolution conductivity can be captured as the size of the coarse block decreases. In addition, the limiting case is that the coarse block size equals the original fine-scale grids, and thus the 'coarse results' and the fine-scale results would be identical. When the upscaling ratio increased to 20, the upscaled results become more different from the fine-scale solutions, which is also reasonable, because more details of the conductivity field are neglected as the upscaling ratio increases. Furthermore, it can also be seen that the upscaled results with the deep learning method can match those from the numerical method well under different upscaling ratios, which demonstrates the ability of the proposed deep learning method to provide approximately the same upscaling accuracy as the numerical method.

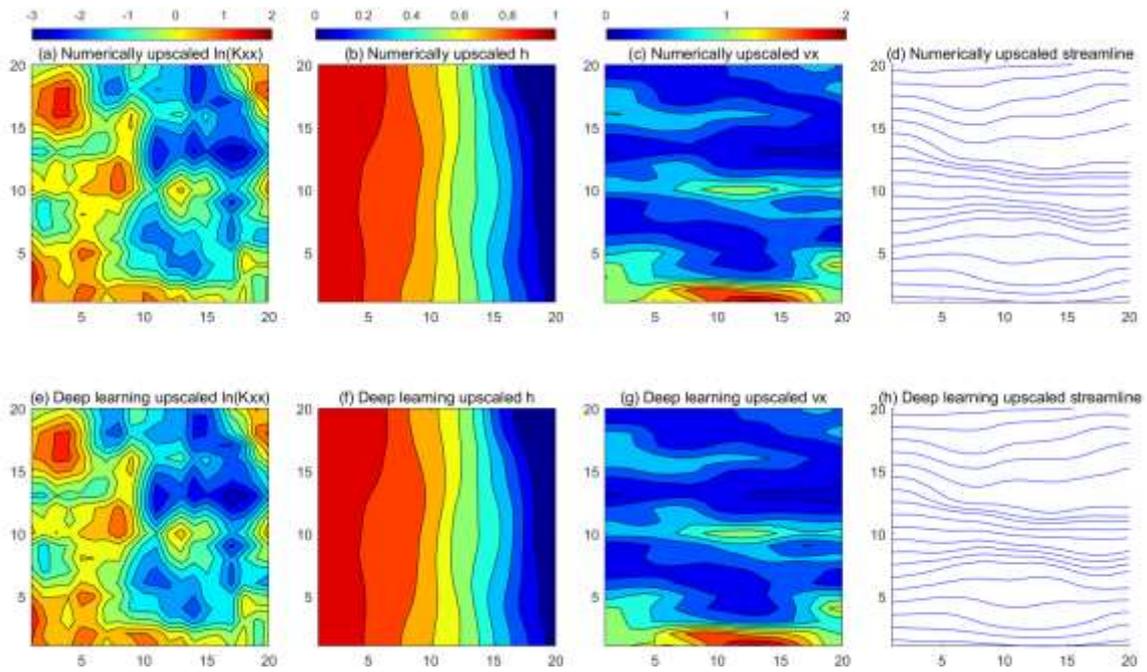

**Figure 11.** Upscaled log-transformed hydraulic conductivity field, hydraulic head, velocity, and streamline with an upscaling ratio of 5.

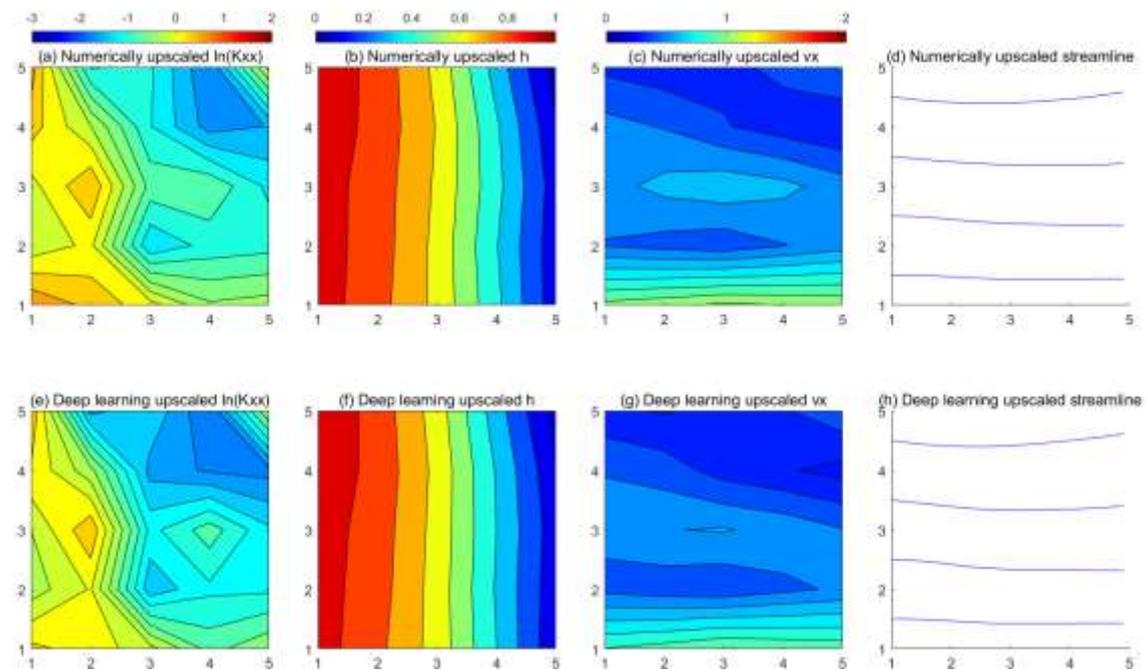

**Figure 12.** Upscaled log-transformed hydraulic conductivity field, hydraulic head, velocity, and streamline with an upscaling ratio of 20.

## 3.2 3D case

In this subsection, the deep learning method is tested in 3D cases, including isotropic and anisotropic scenarios.

### *3.2.1 Isotropic 3D case*

The isotropic scenario of the 3D case is firstly studied in this subsection. The domain size of the fine-scale geologic model is $1200\times 4400\times 700$ [L], which can be divided into $60\times 220\times 35$ grid blocks with each block's size being $20\times 20\times 20$ [L]. The fine-scale geologic model can be upscaled into $12\times 44\times 7$ coarse blocks, with each consisting of $5\times 5\times 5$ fine-scale blocks. The correlation length of the stochastic field in the x-, y-, and z-directions are set to be $\eta_x = 500$ [L], $\eta_y = 1000$ [L], and $\eta_z = 100$ [L], respectively. The mean of the stochastic field $\ln K$ is set to be $E(\ln K) = 0$, and the variance is set to be $\sigma^2_{\ln K} = 2.0$. The exponential covariance function is still utilized:

$$C_{\ln K}(\mathbf{x},\mathbf{x'}) = \sigma^2_{\ln K}\exp(-\frac{|x_1-x_2|}{\eta_x}-\frac{|y_1-y_2|}{\eta_y}-\frac{|z_1-z_2|}{\eta_z}). \qquad (26)$$

The 3D fine-scale hydraulic conductivity realizations can also be generated with KLE (Wang et al., 2021a), and 90% energy is preserved in this case. An example of the generated fine-scale conductivity field is presented in **Figure 13** (a). The hydraulic conductivity field is also assumed to be isotropic in this case, i.e., $Kx = Ky = Kz$, and the anisotropic scenario is investigated in the next subsection.

In the numerical method, the periodic boundary conditions should be generalized into 3D scenarios as follows:

$$\begin{aligned}
H(x,y,z)\big|_{x=0} &= H(x,y,z)\big|_{x=L_x} - \Delta H_1, & v_x(x,y,z)\big|_{x=0} &= v_x(x,y,z)\big|_{x=L_x} \\
H(x,y,z)\big|_{y=0} &= H(x,y,z)\big|_{y=L_y} - \Delta H_2, & v_y(x,y,z)\big|_{y=0} &= v_y(x,y,z)\big|_{y=L_y}, \\
H(x,y,z)\big|_{z=0} &= H(x,y,z)\big|_{z=L_z} - \Delta H_3, & v_z(x,y,z)\big|_{z=0} &= v_z(x,y,z)\big|_{z=L_z}
\end{aligned} \qquad (27)$$

and the physical constraints of the periodic boundary conditions should also be revised accordingly:

$$L_{BC-H}(\theta) = \frac{1}{N_{grid-b}} \frac{1}{N_r} \sum_{i=1}^{N_r} \left\| \left( \hat{H}\big|_{x=0} - \hat{H}\big|_{x=L_x} \right) - \Delta H_1 \right\|_2^2 + \frac{1}{N_{grid-b}} \frac{1}{N_r} \sum_{i=1}^{N_r} \left\| \left( \hat{H}\big|_{y=0} - \hat{H}\big|_{y=L_y} \right) - \Delta H_2 \right\|_2^2$$
$$+ \frac{1}{N_{grid-b}} \frac{1}{N_r} \sum_{i=1}^{N_r} \left\| \left( \hat{H}\big|_{z=0} - \hat{H}\big|_{z=L_z} \right) - \Delta H_3 \right\|_2^2$$

(28)

$$L_{BC-v}(\theta) = \frac{1}{N_{grid-b}} \frac{1}{N_r} \sum_{i=1}^{N_r} \left\| \hat{v}_x\big|_{x=0} - \hat{v}_x\big|_{x=L_x} \right\|_2^2 + \frac{1}{N_{grid-b}} \frac{1}{N_r} \sum_{i=1}^{N_r} \left\| \hat{v}_y\big|_{y=0} - \hat{v}_y\big|_{y=L_y} \right\|_2^2$$
$$+ \frac{1}{N_{grid-b}} \frac{1}{N_r} \sum_{i=1}^{N_r} \left\| \hat{v}_z\big|_{z=0} - \hat{v}_z\big|_{z=L_z} \right\|_2^2$$

(29)

In this case, the hydraulic head differences in the x-, y-, and z-directions are set to be $\Delta H_1 = 1$, $\Delta H_2 = 0$, and $\Delta H_3 = 0$, respectively. In order to construct TgCNN models for the 3D case, 3D convolution modules are adopted. The inputs of the 3D TgCNN are the 3D blocks of the hydraulic conductivity patches, and the outputs are the corresponding hydraulic head solutions for the conductivity patches. The details of the TgCNN structure are listed in **Table 2**. To train the TgCNN model, 10 fine-scale conductivity fields are generated with KLE, each of which can be divided into $12 \times 44 \times 7$ coarse blocks. Therefore, 36,960 coarse blocks are utilized to train the TgCNN model. It is worth mentioning that the 3D TgCNN model is also trained in a 'label-free' manner in this case, because the 36,960 coarse blocks are not solved with the numerical method to provide the hydraulic head labels, and only used to calculate the residuals of the governing equation. The weights $\lambda_{GE}$ and $\lambda_{BC-v}$ are set to be 0.001 in this case, and the rest are still set to be 1. The learning rate is set to be 0.001, and it decays 10% after each 10 epochs. It takes approximately 2128.286 s to train the TgCNN model for 300 training epochs.

Table 2. Architecture of the TgCNN model for the isotropic 3D case.

|  | Layers | Output size | Number of channels |
|---|---|---|---|
| Encoder | Input | 5 × 5 × 5 | 1 |
|  | Convolution (k3s1p1) | 5 × 5 × 5 | 16 |

|         | Activation (Swish)    | $5 \times 5 \times 5$ | 16 |
|         | Convolution (k3s1p1)  | $5 \times 5 \times 5$ | 32 |
|         | Activation (Swish)    | $5 \times 5 \times 5$ | 32 |
|         | Convolution (k3s1p0)  | $3 \times 3 \times 3$ | 64 |
|         | Activation (Swish)    | $3 \times 3 \times 3$ | 64 |
|         | Deconvolution (k3s1p0)| $5 \times 5 \times 5$ | 32 |
|         | Activation (Swish)    | $5 \times 5 \times 5$ | 32 |
| Decoder | Deconvolution (k3s1p0)| $7 \times 7 \times 7$ | 16 |
|         | Activation (Swish)    | $7 \times 7 \times 7$ | 16 |
|         | Deconvolution (k3s1p1)| $7 \times 7 \times 7$ | 1  |
|         | Activation (Swish)    | $7 \times 7 \times 7$ | 1  |

The trained 3D TgCNN model can then be used for efficient upscaling of the 3D fine-scale hydraulic conductivity fields following the workflow in **Figure 3**. The numerical upscaling method is utilized as a reference. The upscaled results of the fine-scale realization shown in **Figure 13** (a) are presented in **Figure 13** (b) and (c). The fine-scale hydraulic head distribution and upscaled results with different methods are shown in **Figure 14**. It can be seen that the upscaled conductivity with the proposed deep learning method is almost the same as that from the numerical method, both of which are similar to the fine-scale conductivity field. The upscaled hydraulic heads with the deep learning method are also similar to those of the numerical references, as well as the fine-scale results. The scatterplots of the upscaled results are presented in **Figure 15**. The deep learning method provides almost identical upscaled results to those of the numerical method. The upscaled velocity in the z-direction with both the deep learning method and the numerical method seems to be worse than those in the x- and y-directions, which may be attributable to the larger heterogeneity in the z-direction. This case demonstrates the feasibility of upscaling for 3D cases with the proposed deep learning method. The more complicated anisotropic 3D case will be investigated in the next subsection.

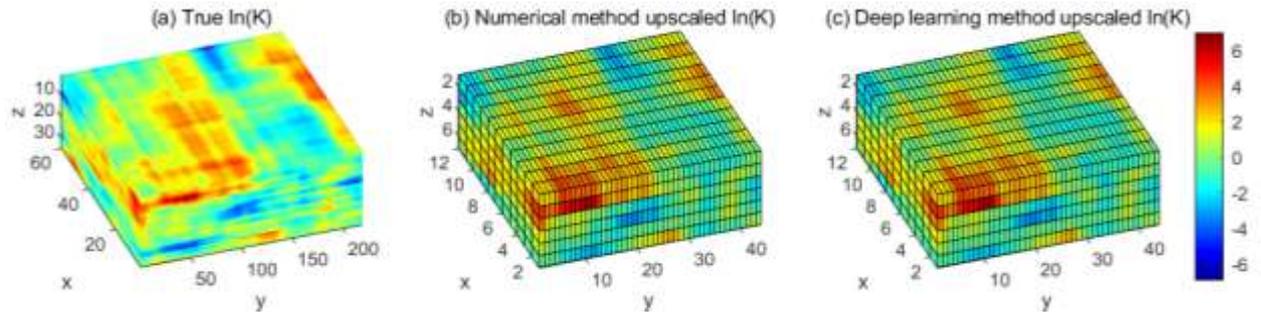

**Figure 13.** Upscaled log-transformed hydraulic conductivity field for the isotropic 3D case.

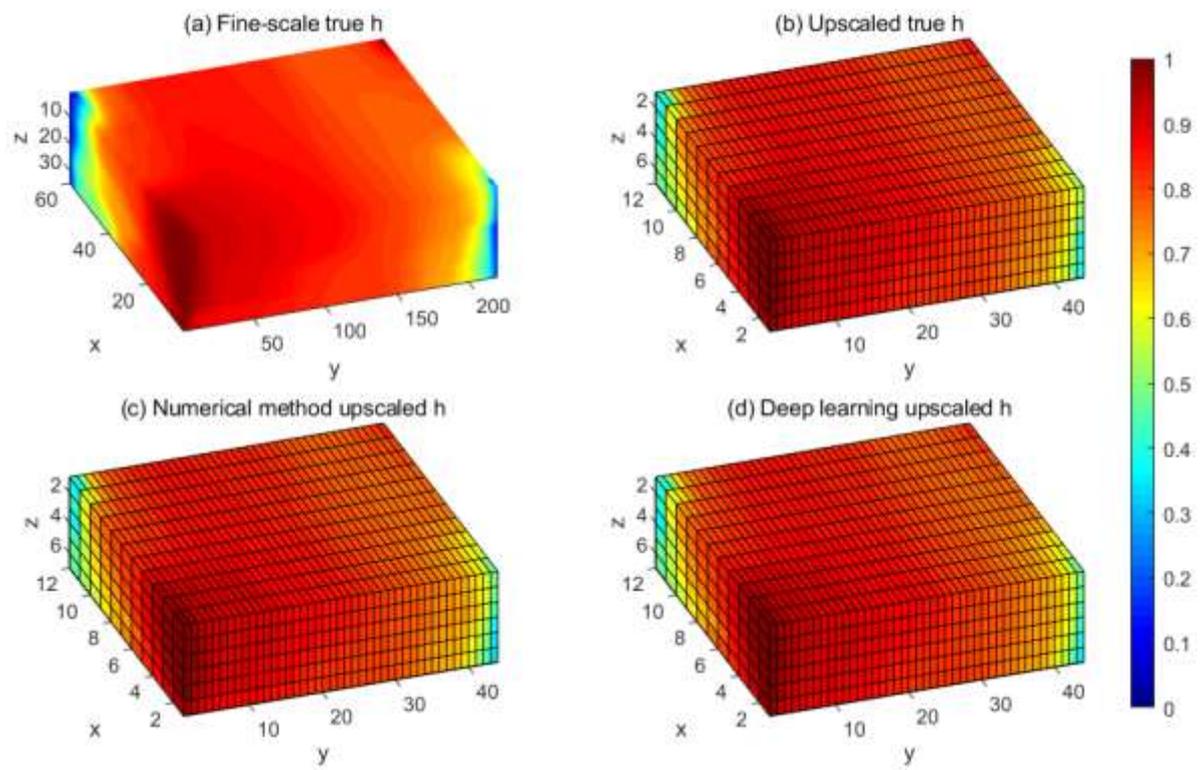

**Figure 14.** Upscaled hydraulic head for the isotropic 3D case.

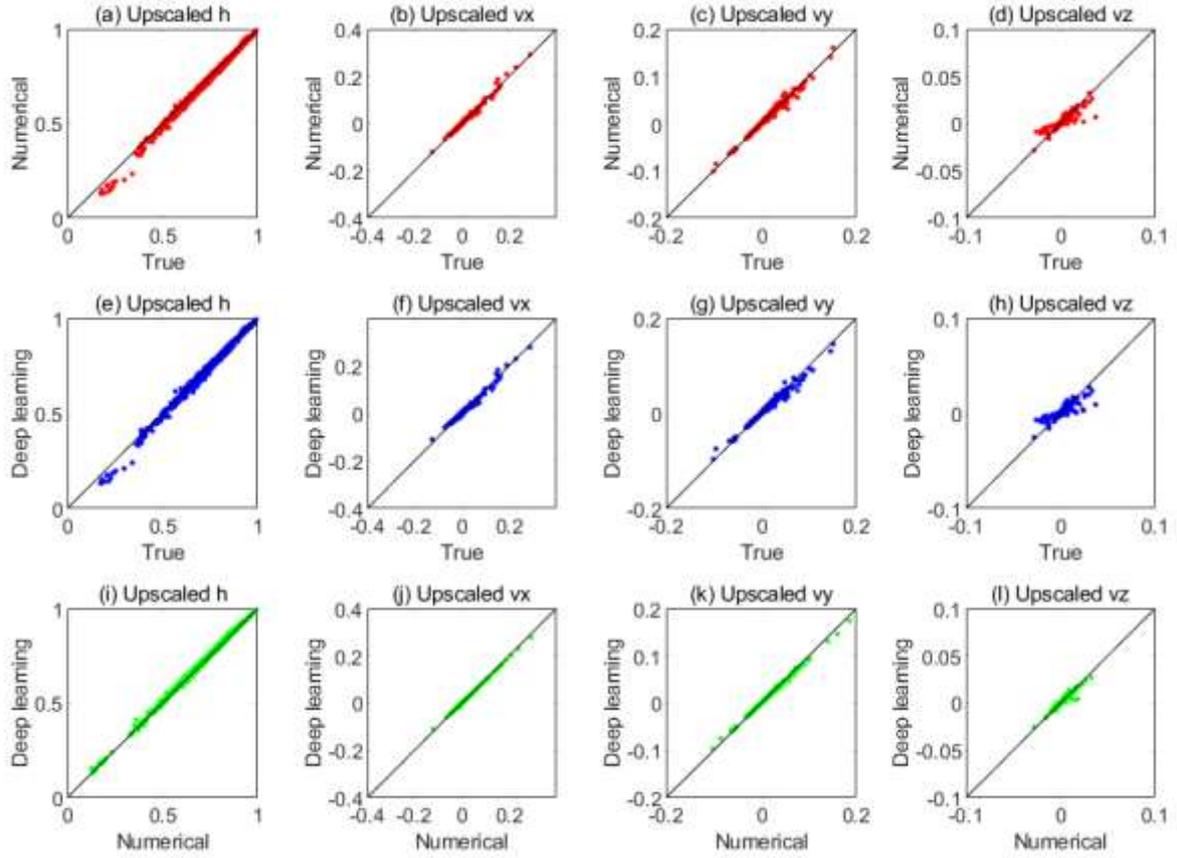

**Figure 15.** Scatterplots of the upscaled results with different methods for the isotropic 3D case.

### 3.2.2 Anisotropic 3D case

In this subsection, a more complicated 3D case is studied, in which the hydraulic conductivity field is anisotropic. The size of the physical domain is set to be $3600\,[L]\times 5000\,[L]\times 1200\,[L]$, and the correlation length of different directions are set as $\eta_x = 1440\,[L]$, $\eta_y = 1000\,[L]$, and $\eta_z = 180\,[L]$, respectively. The statistics, i.e., mean, variance and covariance function, are set to be the same as those in the former case. The domain can be discretized into $180\times 250\times 60$ grid blocks, which is a relatively large-scale model with 2,700,000 grid blocks. The fine-scale model can be coarsened into $36\times 50\times 12$ grid blocks with an upscaling ratio of $5\times 5\times 5$. Furthermore, the most important difference is that the conductivity in the y-direction and the z-direction are assumed to follow $Ky = 0.8Kx$ and $Kz = 0.3Kx$, respectively. The structure of the TgCNN model would be slightly different for this scenario. Considering the anisotropy of the hydraulic conductivity, the inputs of the TgCNN model include the conductivity of x-, y-, and z-directions, i.e., the inputs have three channels, as shown in **Figure 16**.

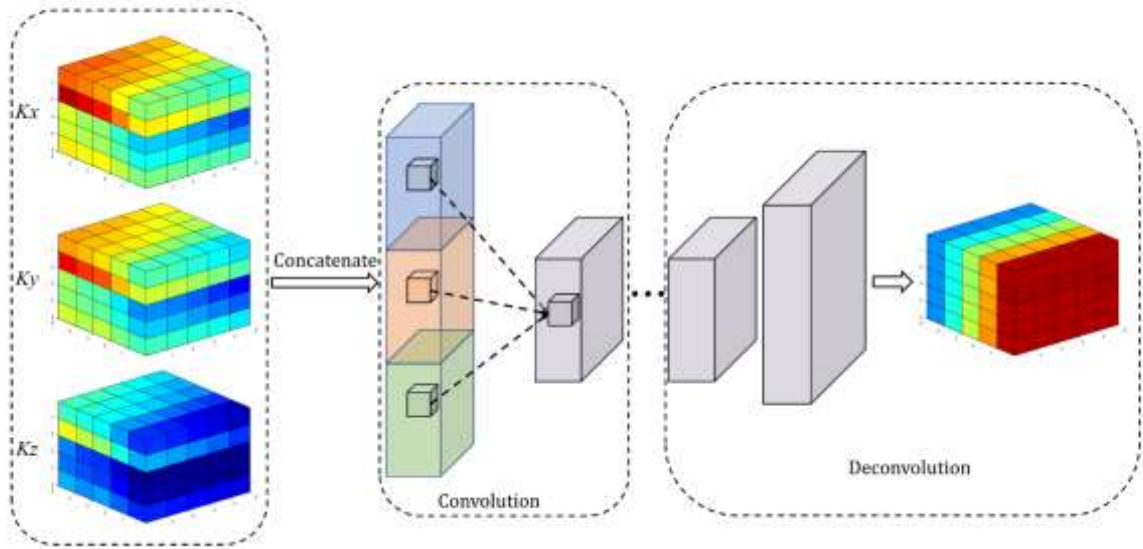

**Figure 16.** The structure of the TgCNN model for anisotropic scenarios.

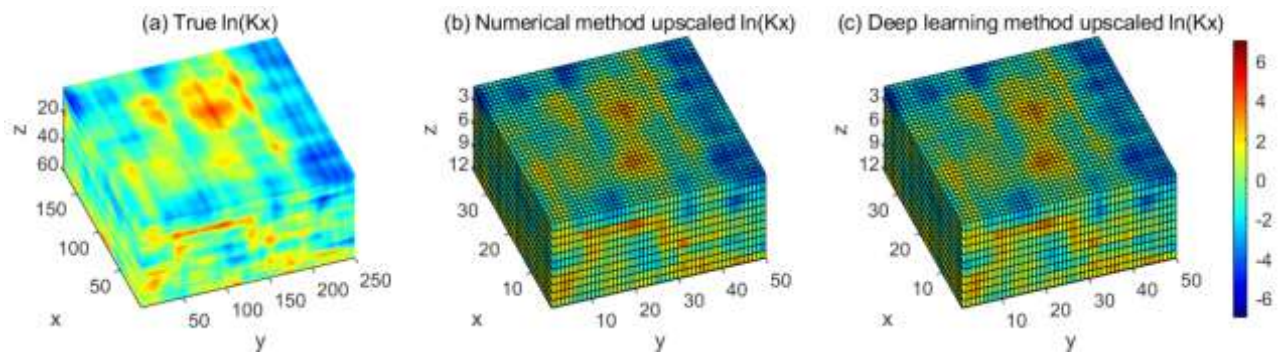

**Figure 17.** Upscaled log-transformed hydraulic conductivity field for the anisotropic 3D case.

To train the TgCNN model for this anisotropic case, a fine-scale realization is generated and divided into 21,600 patches (coarse grid blocks) to impose the physical constraints in the training process. In addition, 3,000 patches of conductivity are numerically solved with periodic conditions to provide the labeled training data, which is a relatively small amount of data considering that each fine-scale model can be divided into 21,600 patches. The network was trained for 200 epochs, and it takes approximately 1277 s to finish the training of the TgCNN model. The trained TgCNN model can then be used for efficient upscaling of the anisotropic fine-scale conductivity realizations.

**Figure 17** (a) presents one realization of fine-scaled $\ln(Kx)$, and the upscaled results with the numerical method and the deep learning method are presented in **Figure 17** (a) and (b), respectively. It is evident that the upscaled $\ln(Kx)$ with the deep learning method can match the results obtained from the numerical method. Moreover, the general pattern of the true fine-scale $\ln(Kx)$ has also been captured, even in the coarse-scale resolution. The upscaled hydraulic conductivity can then be inputted into the simulator to calculate the coarse-scale hydraulic head.

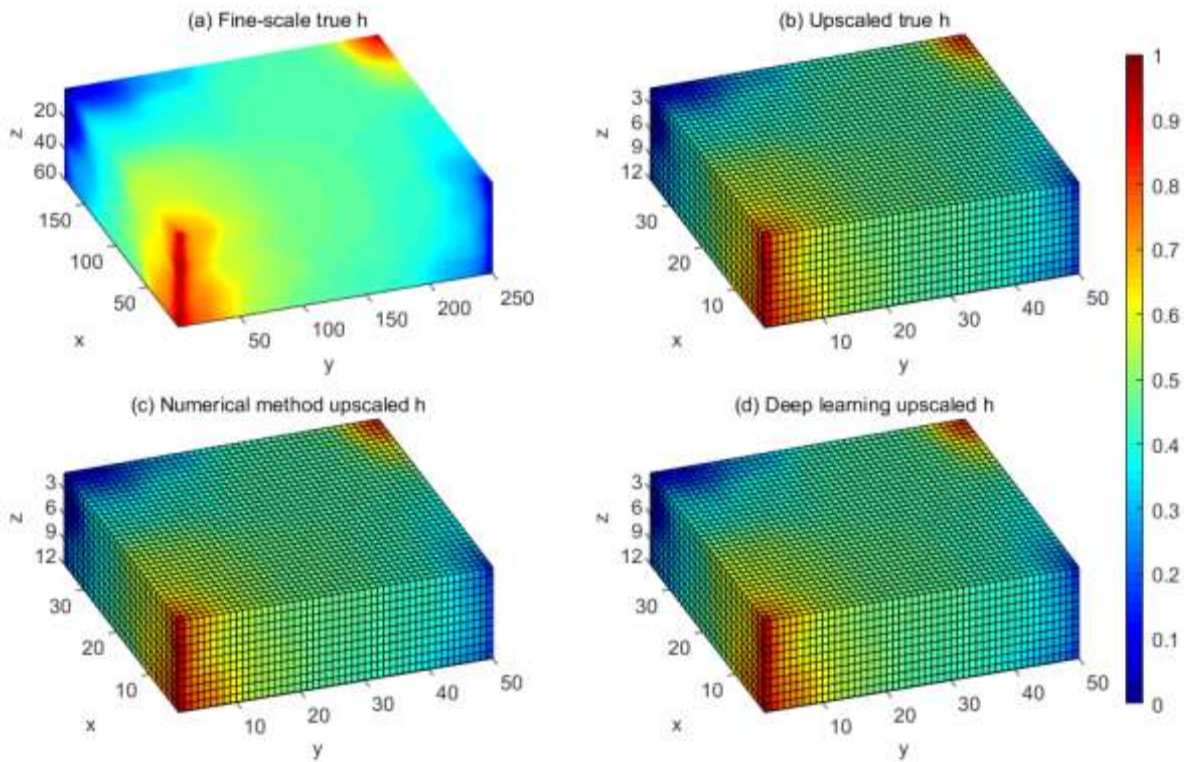

**Figure 18.** Upscaled hydraulic head for the anisotropic 3D case.

**Figure 18** (a) presents the fine-scale hydraulic head of the realization shown in **Figure 17** (a). The true upscaled hydraulic head obtained by averaging the fine-scale results directly is presented in **Figure 18** (b). **Figure 18** (c) and (d) show the hydraulic head obtained by inputting the upscaled conductivity with the numerical method and the deep learning method into the simulator, respectively. It can be seen that the upscaled hydraulic heads with the deep learning method are almost the same as the numerical upscaled results and the benchmark obtained by averaging the fine-scale hydraulic heads. The scatterplots presented in **Figure 19** show the comparison of upscaled results among the benchmark, the numerical method, and the deep learning method. It can be seen that satisfactory accuracy is achieved of the upscaled hydraulic heads and velocity with the deep learning method. Indeed, this case illustrates the effectiveness of the proposed deep learning method for upscaling of anisotropic cases.

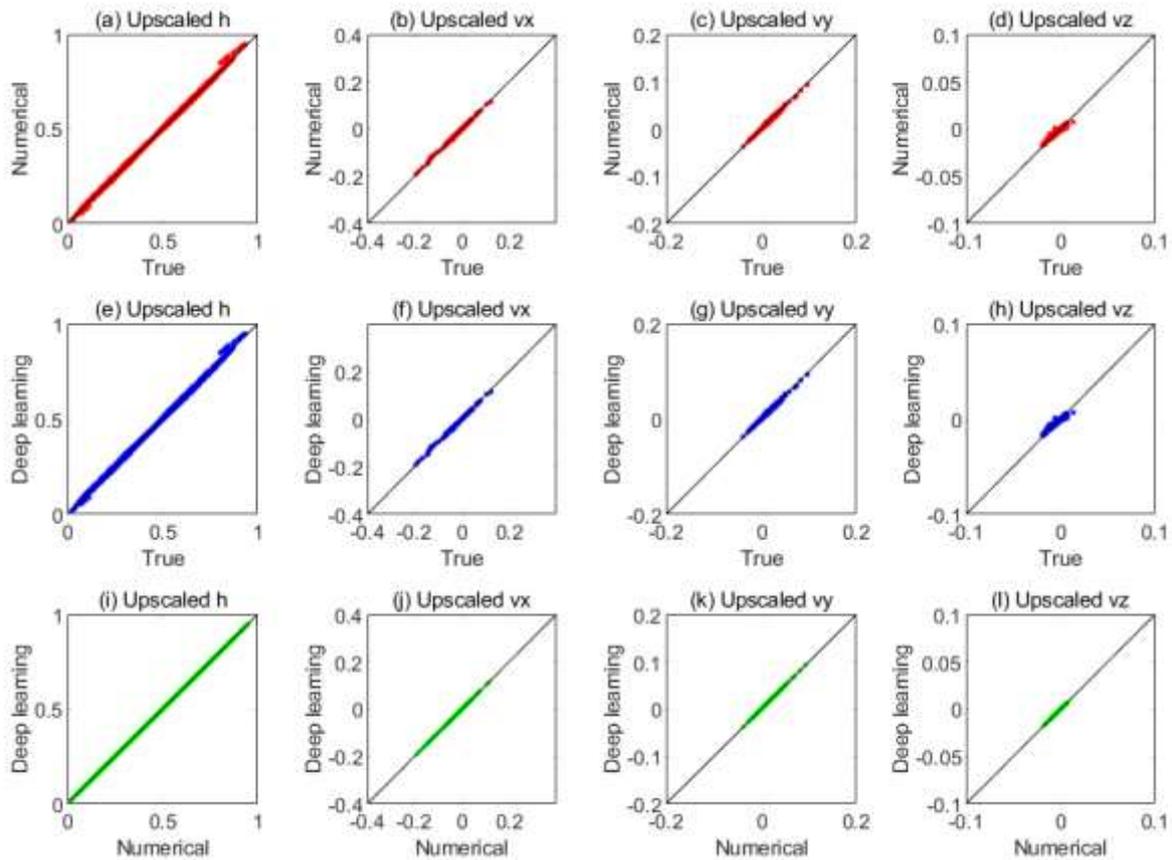

**Figure 19.** Scatterplots of the upscaled results with different methods for the anisotropic 3D case.

The TgCNN model is then utilized for upscaling of 20 fine-scale realizations of hydraulic conductivity fields. The box plots of the $R^2$ score between the upscaled results with the deep learning method and the numerical method are shown in **Figure 20**. One can see that the $R^2$ scores are higher than 0.9 and approach 1, which indicates that the deep learning method can provide statistically equivalent upscaling accuracy to the numerical method. The scatterplots of the upscaled results at five sampled points of the 20 realizations in **Figure 21** also verify this conclusion. Furthermore, the upscaling efficiency of the deep learning method and the numerical method can also be compared in this case. The consumed time for upscaling the hydraulic conductivity fields and solving the flow equations for the 20 fine-scale realizations with different methods are listed in **Table 3**. It can be seen that solving large-scale models (2,700,000 grid blocks) directly without any upscaling process is both computationally expensive and time-consuming. For coarsened models with upscaling methods, however, the time consumed to solve the flow equations can be significantly reduced. It only takes approximately 10 s to solve the 20 upscaled models, which illustrates the effectiveness of the upscaling process. Moreover, the upscaling process can be further accelerated with the deep learning method. It only takes approximately 1 min to upscale the 20 fine-scale models with the trained TgCNN model; whereas, approximately 2 h are required with the numerical method. It is obvious that the deep learning method is superior to the numerical method in terms of efficiency, even when taking training time into account.

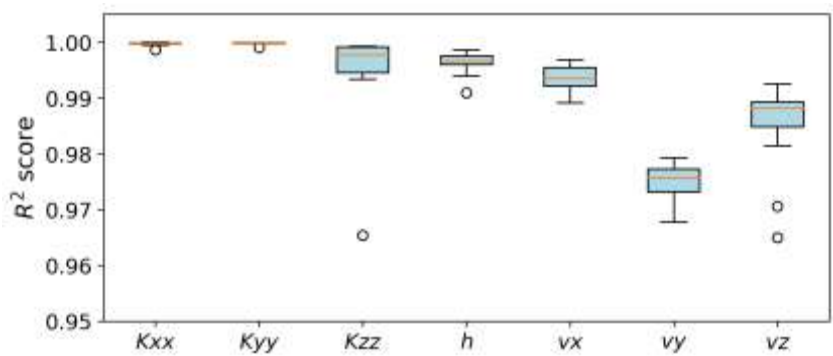

**Figure 20.** Box plots of $R^2$ scores for the upscaled results of 20 fine-scale 3D anisotropic realizations with the deep learning method.

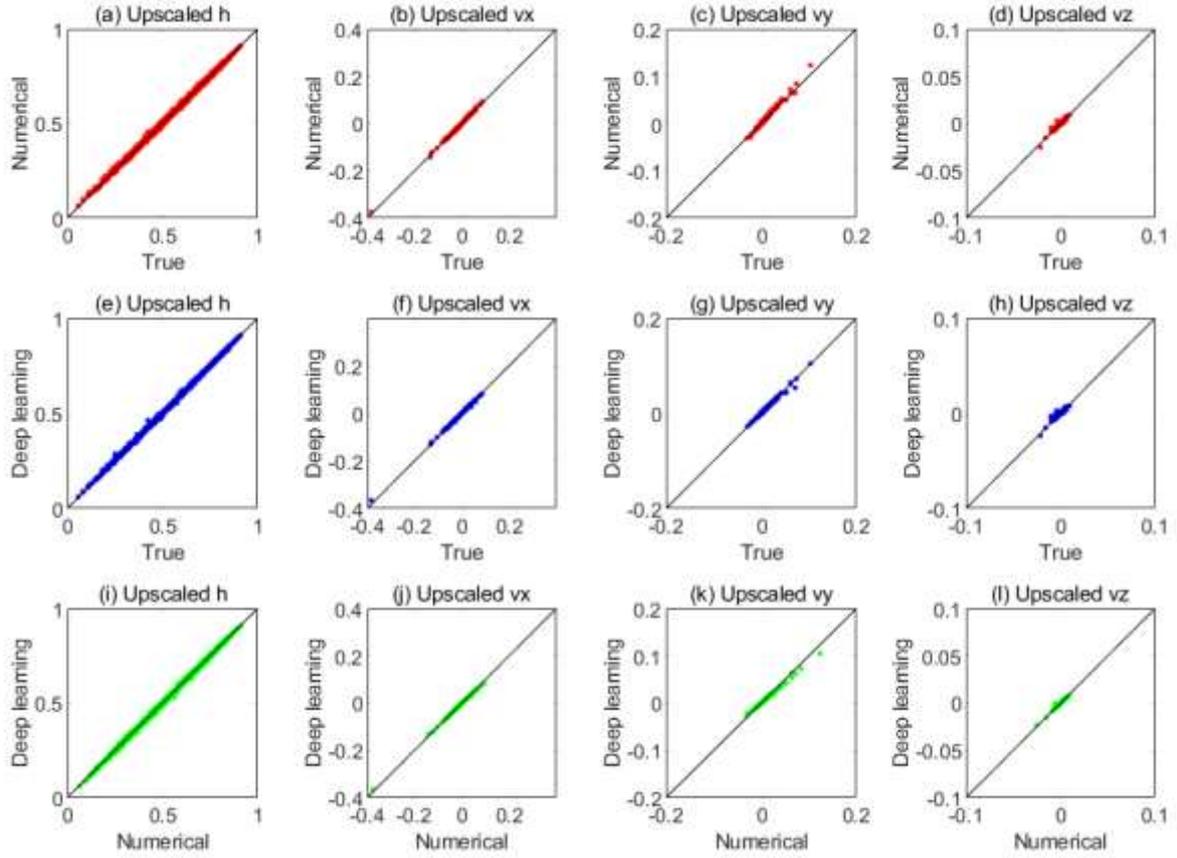

**Figure 21.** Scatterplots of the upscaled results at five sampled points (x=5, y=10, z=2; x=10, y=20, z=4; x=15, y=30, z=6; x=20, y=40, z=8; x=25, y=50, z=10) for the 20 fine-scale realizations.

**Table 3.** Efficiency comparison for solving the 20 fine-scale anisotropic 3D realizations with different methods.

| Method | Training time (s) | Upscaling time (s) | Solving time (s) | Total (s) |
| --- | --- | --- | --- | --- |
| Fine-scale | - | - | 13248.482 | 13248.482 |
| Numerical | - | 6982.053 | 10.678 | 6992.730 |
| Deep learning | 1277.082 | 61.933 | 10.435 | 1349.450 |

## 4 Discussions and Conclusions

A deep-learning-based upscaling method for geologic models was proposed in this work. In the traditional numerical upscaling method, the flow equations need to be solved for each coarse grid block under periodic boundary conditions, which is time-consuming and cumbersome. In this work, the numerical solver of the flow equations for the coarse blocks can be replaced with a trained deep convolutional neural network, which can accurately approximate the relationship between the patches of hydraulic conductivity fields and the hydraulic heads, and thus improve upscaling efficiency. In order to alleviate the dependence on data volume while training deep learning models, the theory-guidance, e.g., governing equations, periodic boundary conditions, etc., can be introduced as prior knowledge to regulate the deep learning models, which is termed the theory-guided convolutional neural network (TgCNN). With the information provided by the physical equations, the TgCNN models can be trained with limited paired data, or even in a label-free manner.

Several cases are introduced to test the performance of the proposed deep-learning-based upscaling method, including 2D and 3D cases. The trained TgCNN models can provide accurate predictions of the hydraulic heads for different patches of conductivity fields, which can be further used for efficient upscaling of fine-scale geologic models. The numerical method was utilized as a reference. The results demonstrate that the deep learning method can provide equivalent upscaling accuracy to the numerical method, whether in 2D/3D cases, or isotropic/anisotropic cases. Furthermore, the upscaled hydraulic heads and velocity can match the directly averaged results of the original fine-scale models well. Although only steady cases are introduced to verify the performance of the proposed method, it can be utilized for upscaling of transient cases, because the permeability fields usually do not change over time.

The efficiency of upscaling can be improved significantly with the deep learning method. With the trained deep learning models, the time-consuming numerical solving process can be bypassed, and the forward calculation of deep learning models is much faster than the numerical solvers. Even taking the training time of deep learning models into consideration, the deep learning method still offers more advantages regarding efficiency than the numerical method. Furthermore, for certain large-scale geologic models or tasks that require a large number of realizations (e.g., uncertainty quantification, data assimilation), the superiority of the proposed deep learning upscaling method would be much more obvious.

**Acknowledgements**

This work is partially funded by the Shenzhen Key Laboratory of Natural Gas Hydrates (Grant No. ZDSYS20200421111201738) and the SUSTech - Qingdao New Energy Technology Research Institute.